\pdfoutput=1

\documentclass[11pt]{article}

\usepackage{booktabs, multirow} %
\usepackage{soul}%
\usepackage[table]{xcolor} %

\usepackage[]{EMNLP2023}

\usepackage{times}
\usepackage{latexsym}

\usepackage[T1]{fontenc}

\usepackage[utf8]{inputenc}

\usepackage{graphicx}

\usepackage{microtype}

\usepackage{inconsolata}

\title{Increasing The Performance of Cognitively Inspired Data-Efficient Language Models via Implicit Structure Building}

\author{Omar Momen\textmd{,} David Arps \textmd{and} Laura Kallmeyer\\
  Heinrich Heine University \\
  Düsseldorf, Germany \\
  \texttt{\{omar.hassan,david.arps,laura.kallmeyer\}@hhu.de} \\ 
}

\begin{document}
\maketitle

\begin{abstract}
In this paper, we describe our submission to the BabyLM Challenge 2023 shared task on data-efficient language model (LM) pretraining \cite{warstadt-et-al-2023-babylm}. 
We train transformer-based masked language models that incorporate unsupervised predictions about hierarchical sentence structure into the model architecture. 
Concretely, we use the Structformer architecture \cite{shen-etal-2021-structformer} and variants thereof. 
StructFormer models have been shown to perform well on unsupervised syntactic induction based on limited pretraining data and to yield performance improvements over a vanilla transformer architecture \citep{shen-etal-2021-structformer}.
Evaluation of our models on 39 tasks provided by the BabyLM challenge shows promising improvements of models that integrate a hierarchical bias into the architecture at some particular tasks, 
even though they fail to consistently outperform the baseline model on all tasks.\footnote{Implementation and models checkpoints can be found here: \url{https://github.com/OmarMomen14/structformer-babylm}}

\end{abstract}

\section{Introduction}

Transformer-based Language Model (LM) performance is heavily influenced by three scaling factors: the number of model parameters, the pretraining dataset size, and the amount of computing. For optimal performance, all three factors must be simultaneously scaled up \cite{kaplan2020scaling}. This scaling law has introduced several challenges in advancing research on neural language modeling. One major obstacle lies in the unequal distribution of resources across languages. Consequently, the current approach of transformer-based models falls short of achieving equally high-performance levels for models dedicated to different languages \cite{Choudhury_Deshpande_2021}.

Moreover, we see a considerable difference when comparing the way %
 LMs learn how humans acquire language. One difference concerns the data that is input to learning:  LMs such as BERT \cite{devlin-etal-2019-bert}, RoBERTa \cite{liu2019roberta} or GPT-3 \cite{brown2020language} are exposed to billions of tokens during training, far surpassing what an individual human is exposed to when learning a language \cite{warstadt2022artificial}. This fundamental discrepancy raises important considerations when drawing parallels between language learning in machines and humans.

To improve the data-efficiency of LMs, one %
direction is to adapt the model architecture. An effective approach in this endeavor involves incorporating an inductive bias into the models' architectures, which could potentially facilitate acquiring more knowledge from the same amount of data compared to standard models. However, the specific type of inductive bias to be added is still under exploration. Recently, there have been efforts to investigate the use of syntactic hierarchical inductive biases as a potential improvement \cite{mulligan-etal-2021-structure,papadimitriou2023pretrain}.\footnote{Note that we don't want to claim that humans integrate such an inductive bias and therefore can learn language with less data, compared to large LMs.}

One of these potential solutions is the StructFormer architecture \cite{shen-etal-2021-structformer}, a transformer that is trained on the masked language modeling task. An additional convolutional neural network (CNN) component produces unlabeled dependency and constituency trees as a byproduct and influences the self-attention mechanism of the transformer layers. 
The model has obtained demonstrated competitive results in structure induction evaluations and a decrease in perplexity over a vanilla transformer baseline \cite{vaswani2017attention}. However, it is an open question whether the inductive bias learned in this architecture enhances performance on downstream NLP tasks.

We pretrain the StructFormer architecture on a dataset from a different domain that had not been tested on that model before. Moreover, we use a more sophisticated tokenizer in comparison to the most frequent words dictionary used to train the models in the original experiment. 
Additionally, we modify the model architecture %
to investigate whether injecting a hierarchical bias in the middle layers of the transformer architecture (rather than after the embedding layer) leads to improved downstream performance.
Eventually, we evaluate seven model variants through the evaluation pipeline of the shared task and submit our best-performing model to the shared task challenge.

\subsection{The BabyLM Challenge}
The BabyLM Challenge is a shared task with the aim of data-efficient language modeling for English. Participants pretrain a LM  from scratch on data that corresponds to the amount of linguistic data available to a child. The task is a great setting for conducting our experiments. It %
 provides us with a pretraining dataset, a thorough evaluation pipeline, and, furthermore, %
  an environment where we can compare our models' performance to other interesting architectures from the systems participating in the shared task. %

  \paragraph{Dataset} The shared task is conducted in two tracks with different dataset sizes: a 100M words corpus, and a 10M words corpus as a sample of the larger corpus. 
  The size is inspired by the assumption that children are exposed to 2M-7M words per year \cite{gilkerson2017mapping}. 
  To account for the fact that children mostly interact with spoken rather than written language data, the datasets include a high proportion of transcribed data from different domains. 
  For more details regarding the source domains, please refer to \citet{warstadt-et-al-2023-babylm}. 

  \paragraph{Evaluation} A thorough evaluation pipeline that comprises 39 different tasks is used to evaluate every model participating in the shared task. These tasks are supposed to represent a model's performance with respect to efficiency and applied NLP, as well as cognitive science, and linguistics. A group of 17 tasks, named \textit{BLiMP} \cite{warstadt-etal-2020-blimp-benchmark} are performed via zero-shot predictions, while the other two groups of tasks; \textit{SuperGLUE} \cite[11 tasks, ][]{10.5555/3454287.3454581} and \textit{MSGS} \cite[11 tasks,][]{warstadt-etal-2020-learning} need finetuning of the submitted models for classification. Refer to Appendix \ref{appendix: evaluation_metrics} for the complete list of tasks.

\section{Language Modeling and Hierarchical Information}

Transformer LMs use syntactic information in their predictions. 
This has been shown by work on interpreting their internal representations as well as by investigating the grammatical correctness of their predictions \cite{mahowald-etal-2023-dissociating,kulmizev-nivre-2022-schroedingers}.
However, the vanilla transformer architecture that underlies both encoder and decoder-based LMs does not encode hierarchical information explicitly. Rather, objectives such as masked language modeling and next-token prediction are based on linear relationships between tokens. 
This has inspired two lines of work that incorporate hierarchical knowledge into LMs.
The first group of papers introduces models in which the training objective involves syntactic labels explicitly \citep[e.g.][]{dyer-etal-2016-recurrent,sartran-etal-2022-transformer}, 
The second group introduces models in which hierarchical information is encoded implicitly as a byproduct of a language modeling task 
\citep{shen2018neural,shen-etal-2021-structformer,li-etal-2019-imitation,kim-etal-2019-unsupervised,choi-etal-2018-learning,williams-etal-2018-latent}. 
We consider the second group of models more relevant for this shared task since it allows us to train models with a hierarchical architecture bias on raw text data. 
In particular, we use the StructFormer model \cite{shen-etal-2021-structformer},  a transformer in which one architecture component, the parser network, predicts the position of each token in the hierarchical structure of the sentence. The prediction of the parser network puts soft constraints on the attention mask of the transformer layers. The model is pretrained on the masked language modeling task, and we view two experimental contributions of \citet{shen-etal-2021-structformer} as most relevant for using this model: First, they show that a StructFormer achieves lower perplexity on limited training data than a transformer that replaces the parser network with standard self-attention. 
Second, the induced hierarchical structure corresponds to unlabeled dependency trees. Concretely, evaluation on the Penn Treebank (PTB) shows that 61.6\% of the undirected dependency edges are recovered. 
We further implement a variant of the model in which the parser network predicts hierarchical information based on hidden states that are contextualized with classical transformer layers, rather than using uncontextualized token embeddings as direct input to the parser network (Sec.~\ref{sec:model-architecture}).

\section{Experiment}

This section introduces the objectives of our experiment%
, a description of the model architectures, and the technical aspects of the pretraining and evaluation process.

\subsection{Objectives}\label{sec:objectives}

In this work, we aim to validate the claim that the performance of LMs, in particular on syntax-sensitive tasks, can be improved %
 through the implicit integration of an inductive bias into the model's architecture that yields a hierarchical structure of the tokens. %
  Concretely, we conduct experiments towards pursuing the following %
   three primary objectives: %

\begin{enumerate}
    
    \item \label{objective_1} Assess the robustness of the finding that LM performance is enhanced through the utilization of a linguistically informed model architecture \cite{shen-etal-2021-structformer}.
    
    \item \label{objective_2} Investigate whether the claim that transformer architectures better represent syntactic information in their middle attention layers is supported in a practical use case \cite{vig-belinkov-2019-analyzing,arps-etal-2022-probing,muller-eberstein-etal-2022-probing}.
    \item \label{objective_3} Develop models that surpass %
     the performance of the baseline models offered by the organizers of the shared task.

\end{enumerate}

\subsection{Methodology}\label{sec:exp-methodology}

In order to address the questions posed by the experiment's objectives, we train a tokenizer, develop several model variants, and perform iterations of model pretraining, finetuning, and evaluation.
Due to limited resources, we only conducted our experiments on the 10M words dataset. 
Furthermore, from the model architectures provided by the shared task, we chose the encoder-type models due to their adaptability for integrating a hierarchical bias in the model architecture. 

\subsubsection{Tokenizer}

We use the same tokenizer across all variations of our models. Specifically, we train a Byte Pair Encoding (BPE) tokenizer \cite{sennrich-etal-2016-neural,gage-1994-bpe} %
from scratch on the 10M BabyLM corpus. %
Since BPE tokenizers require specifying the vocabulary size as a hyperparameter before training on the corpus, we carefully determined an appropriate size. Our goal was to obtain a tokenizer that accurately represents tokens in our relatively small dataset while adhering to best practices for LMs.
To achieve this, we train the tokenizer on the same corpus with different vocabulary sizes. We then observed the resulting vocabularies and identified the least frequent tokens within each (Table \ref{tokenizers}).

Based on our analysis, a vocabulary size of 32K tokens provides a fair representation relative to the corpus size for the least frequent tokens. Additionally, \citet{geiping2022cramming} found that a BPE tokenizer with 32K tokens yielded the best results.

\begin{table}[!t]\centering
\scriptsize
\begin{tabular}{lrrr}\toprule
Vocabulary Size &Least Frequent Tokens &Frequency \\\midrule
8K &sought, arts, stolen, ATOR &230 \\
10k &accounts, seated, lemn, feathers &165 \\
12k &sailors, goss, reun, irlines &126 \\
16k &sophisticated, olleyball, AMES, poorly &80 \\
32k &jets, estus, iesselin, UCLA, mannik &26 \\
\bottomrule
\end{tabular}
\caption{Tokenizer Vocabulary Size Experiments}\label{tokenizers}
\end{table}

\subsubsection{Baseline model}

To achieve objective \ref{objective_1}, we pretrained a standard transformer architecture that we call \emph{transformer-base}, using our custom-trained tokenizer and following the same model and training hyperparameters to minimize any effects due to uncontrolled variables.

\subsubsection{Hyperparameters}

Due to resource limitations, and to assure fair comparisons between models, we use one set of pretraining and finetuning hyperparameters: We chose the default hyperparameters settings that were used to pretrain the shared task baseline models \cite{warstadt-et-al-2023-babylm}.
In order to speed up the evaluation of finetuning tasks, we made modifications to the finetuning hyperparameters that were used to evaluate the baseline models. Our main hyperparameters are reported in Appendix \ref{appendix: hyperparams}.
We pretrain all models with the same batch size and the same number of steps. 
We use the training pipeline that \citet{warstadt-et-al-2023-babylm} introduced to train their baseline modes to minimize any effects due to uncontrolled variables.

However, one variable that could not be fixed during the experiment is the number of trainable parameters in each model. When adding a convolution parser network to a particular model, the increase in the number of parameters in that model is inevitable (parameter counts are listed in Appendix \ref{appendix: hyperparams}). We are aware that this can have misleading effects on the results and conclusions, however, we still think that the experiment in its current setting can show interesting behaviors that may encourage further investigation in a fully controlled experiment.

\subsubsection{Model Architectures}\label{sec:model-architecture}

We develop two primary variants of model architectures for our experiment.

\begin{figure}[!h]
    \centering
    \includegraphics[width=7.2cm]{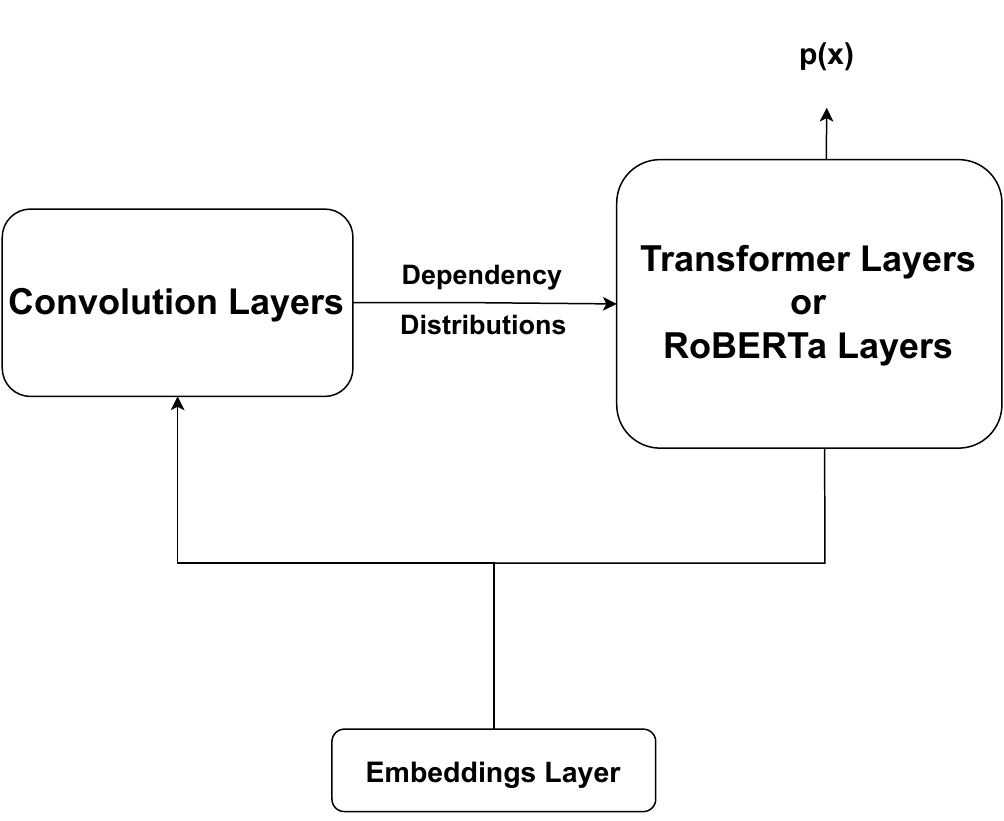}
    \caption{StructFormer and StructRoBERTa Architectures (\(s_1\))}
    \label{fig:structformer}
\end{figure}

\paragraph{StructFormer} This variant (Figure \ref{fig:structformer}) closely follows the architecture in \citet{shen-etal-2021-structformer}. In brief, it incorporates a parser network that consists of 4 convolution layers. The input to the parser network is token embeddings, and the output is probability distributions for dependencies between tokens. These distributions are then integrated into the multi-head self-attention mechanism of a standard transformer model. For a complete description of the architecture, we refer readers to \citet{shen-etal-2021-structformer}. 
We name models of this variant by the prefix \textit{structformer}. 

\paragraph{StructRoBERTa} The second variant (Figure \ref{fig:structformer}) is similar to the StructFormer, but instead of employing a standard transformer, it utilizes a base RoBERTa encoder \cite{liu2019roberta}. We modify the HuggingFace \cite{wolf2020huggingfaces} implementation, which has a few differences from the vanilla transformer implementation, mainly adding normalization and dropout layers after the embeddings layer, and also adding an additional intermediate block within each layer. The models following this architecture will be identified with the prefix \textit{structroberta}.

\paragraph{Vanilla transformer} For transformers without parser networks, we reuse the implementation by \citet{shen-etal-2021-structformer} which follows the standard transformer introduced by \citet{vaswani2017attention}, except that a layer normalization is added in front of each layer.

\begin{figure}[!h]
    \centering
    \includegraphics[height=10cm]{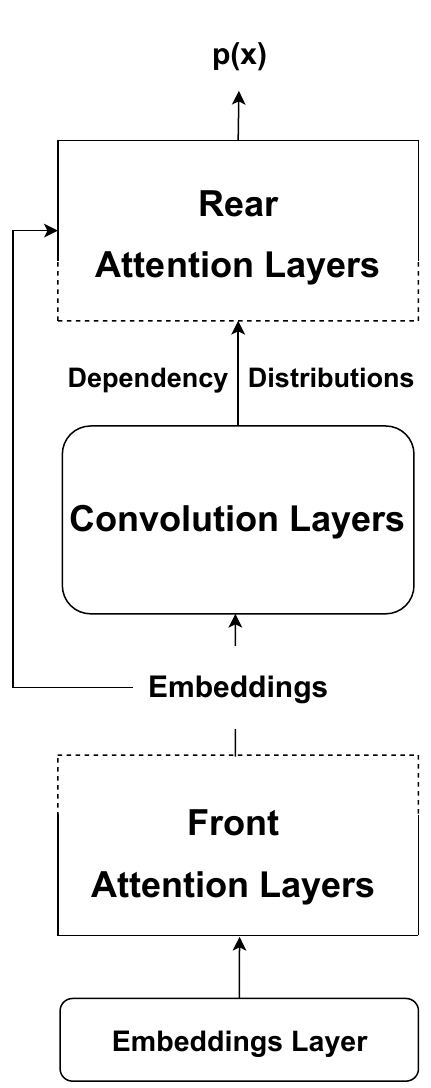}
    \caption{In-between Parser Architectures (\(s_2\)), dotted lines indicate intervening the encoder layers at two positions, where the parser network connects the two split parts of the encoder }
    \label{fig:in-between_parser}
\end{figure}

\paragraph{Variants} Subsequently, for each of the main variants, \textit{structformer} and \textit{structroberta}, we create two sub-variants to explore a different placement of the parser network within the architecture (Figure \ref{fig:in-between_parser}). This decision is based on insights from previous experiments, which indicate that syntactic information tends to be better represented in the middle layers of the transformer \citep{liu-etal-2019-linguistic,vig-belinkov-2019-analyzing,arps-etal-2022-probing}.

In our approach, we divide the initial \(n_{context}\) layers of either the transformer or RoBERTa component in \textit{structformer} or \textit{structroberta} respectively. We label these \(n_{context}\) layers as the Front Attention Layers, while the remaining attention layers are labeled as Rear Attention Layers. The input embeddings pass through the Front component, generating embeddings that are subsequently fed into the parser network. The parser network, in turn, outputs dependency distributions that are integrated into the Rear component of the architecture.

To distinguish between the two sub-variants, we append the suffix \(s_1\) to models with the parser network before the attention layers (Figure \ref{fig:structformer}), and the suffix \(s_2\) to models with the parser network in-between the middle attention layers (Figure \ref{fig:in-between_parser}).

To achieve objective \ref{objective_3}, we introduce two additional models, \textit{structroberta}\(_{s1'}\) and \textit{structroberta}\(_{s2'}\),
to enhance the evaluation scores so we could submit the best attainable results to the shared task. These two models are basically an upgrade in the number of convolution layers (from 4 to 6) of the parser network in \textit{structroberta}\(_{s1}\) and \textit{structroberta}\(_{s2}\) respectively.

\section{Results}

After completing the pretraining process of the 7 investigated models, a comprehensive linguistic evaluation is conducted for the seven models under study. The shared task evaluation pipeline is used for this purpose. Detailed evaluation results are presented in Tables \ref{ppl} \ref{blimp_1}, \ref{glue_1}, and \ref{msgs_1}.
We compare the scores of the following models: \textit{transformer-base} (TF\(_{base}\)), \textit{structformer\(_{s1}\)} (SF\(_{s1}\)), \textit{structformer\(_{s2}\)} (SF\(_{s2}\)), \textit{structroberta\(_{s1}\)} (SR\(_{s1}\)), \textit{structroberta\(_{s2}\)} (SR\(_{s2}\)), \textit{structroberta\(_{s1'}\)} (SR\(_{s1'}\)) and \textit{structroberta\(_{s2'}\)} (SR\(_{s2'}\)). We are particularly interested in assessing to which extent the introduction of a hierarchical bias improves a model's performance on a specific task. Therefore, in addition to the scores of the individual models, we also report the %
 differences in scores as follows:
\begin{itemize}
\item \(\Delta_{SF_{s1}} = Score(SF_{s1}) - Score(TF_{base})\)
\vspace{-.5em}
\item \(\Delta_{SF_{s2}} = Score(SF_{s2}) - Score(TF_{base})\)
\vspace{-.5em}
\item \(\Delta_{SR_{s12}} = Score(SR_{s1}) - Score(SR_{s2})\)
\vspace{-.5em}
\item \(\Delta_{SR_{s1'}} = Score(SR_{s1'}) - Score(SR_{s1})\)
\vspace{-.5em}
\item \(\Delta_{SR_{s2'}} = Score(SR_{s2'}) - Score(SR_{s2})\)
\end{itemize}

All numerical values in the result tables are measures of accuracy unless explicitly stated otherwise.

\subsection{Pseudo-perplexity}

We report the corpus-level pseudo-perplexity \cite[$PPPL$,][]{salazar-etal-2020-masked} on the test split of the BabyLM shared task dataset\footnote{We use \citet{kauf-ivanova-2023-better}'s implementation for computing $PPPL$ scores and remove the 100 longest sentences from the dataset to reduce the computation time.} (Table \ref{ppl}). 
$PPPL$ is computed by masking out each token in turn and collecting the log-likelihoods. 
This evaluation contributes to objective \ref{objective_1} in our experiment. 
\citet{shen-etal-2021-structformer} found that \textit{structformer} models incorporating hierarchical inductive bias achieve lower 
$PPPL$ than their baseline \textit{transformer} model.
We want
to assess this finding on the BabyLM dataset and using our custom-trained tokenizer. 
SF\(_{s1}\) shows lower $PPPL$ compared to TF\(_{base}\), which follows the previous findings. However, the model with a parser network within the middle layers shows a higher $PPPL$ than the baseline TF\(_{base}\). 
The addition of more convolution layers at the parser network shows an improvement at SR\(_{s2'}\) but surprisingly shows a deterioration at SR\(_{s1'}\).

\begin{table*}[!htp]\centering
\scriptsize
\begin{tabular}{lrrr|rrrr}
\toprule
& 
& Set A
&
& 
& Set B
& 
&
\\\midrule

& \textbf{TF\(_{base}\)} 
& \textbf{SF\(_{s1}\)} 
& \textbf{SF\(_{s2}\)} 
& \textbf{SR\(_{s1}\)} 
& \textbf{SR\(_{s2}\)} 
&\textbf{SR\(_{s1'}\)} 
&\textbf{SR\(_{s2'}\)} 
\\\midrule

Perplexity&32.84 &26.48 &38.26 &\textbf{21.15} &23.15 &37.11 &22.48 \\
\bottomrule
\end{tabular}
\caption{Perplexity Results}\label{ppl}
\end{table*}

\subsection{BLiMP}

BLiMP is a challenging benchmark comprising a set of tests designed to evaluate the linguistic knowledge of LMs %
 with a specific focus on linguistic phenomena encompassing syntax, morphology, and semantics \cite{warstadt-etal-2020-blimp-benchmark}. Originally, the benchmark consisted of 12 tasks (see Appendix \ref{appendix: evaluation_metrics}). Additionally, in the shared task \cite{warstadt-et-al-2023-babylm}, 5 more tasks were added to BLiMP as held-out tasks, aiming to assess the generalization capabilities of the submitted models. The random chance accuracy for all original BLiMP tasks is 50, while chance was not reported for the additional 5 supplement tasks. %

\begin{table*}[tp]\centering
\scriptsize
\begin{tabular}{p{2.3cm}rrrrrrrrrrrr}
\toprule
& 
\multicolumn{5}{c}{\textbf{Set A}}    
& &
\multicolumn{5}{c}{\textbf{Set B}}     
& 
\\\midrule

& \textbf{TF\(_{base}\)} 
& \textbf{SF\(_{s1}\)} 
& \textbf{SF\(_{s2}\)} 
&\cellcolor[HTML]{E9E9E9} \textbf{\(\Delta_{SF_{s1}}\) }
&\cellcolor[HTML]{E9E9E9} \textbf{\(\Delta_{SF_{s2}}\) } 
& \textbf{SR\(_{s1}\)} 
& \textbf{SR\(_{s2}\)} 
&\textbf{SR\(_{s1'}\)} 
&\textbf{SR\(_{s2'}\)} 
&\cellcolor[HTML]{E9E9E9}\textbf{ \(\Delta_{SR_{s12}} \) } 
&\cellcolor[HTML]{E9E9E9}\textbf{ \(\Delta_{SR_{s1'}} \) } 
&\cellcolor[HTML]{E9E9E9}\textbf{ \(\Delta_{SR_{s2'}} \) } 
\\\midrule

Anaphor Agreement &88 &88 &74 &\cellcolor[HTML]{E9E9E9}0 &\cellcolor[HTML]{E9E9E9}-14 &89 &87 &90 &87 &\cellcolor[HTML]{E9E9E9}-2 &\cellcolor[HTML]{E9E9E9}1 &\cellcolor[HTML]{E9E9E9}0 \\
Argument Structure &68 &69 &68 &\cellcolor[HTML]{E9E9E9}1 &\cellcolor[HTML]{E9E9E9}0 &69 &72 &73 &68 &\cellcolor[HTML]{E9E9E9}3 &\cellcolor[HTML]{E9E9E9}4 &\cellcolor[HTML]{E9E9E9}-4 \\
Binding &68 &68 &66 &\cellcolor[HTML]{E9E9E9}0 &\cellcolor[HTML]{E9E9E9}-2 &72 &70 &70 &67 &\cellcolor[HTML]{E9E9E9}-2 &\cellcolor[HTML]{E9E9E9}-2 &\cellcolor[HTML]{E9E9E9}-3 \\
Control Raising &66 &66 &64 &\cellcolor[HTML]{E9E9E9}0 &\cellcolor[HTML]{E9E9E9}-2 &69 &70 &68 &63 &\cellcolor[HTML]{E9E9E9}1 &\cellcolor[HTML]{E9E9E9}-1 &\cellcolor[HTML]{E9E9E9}-7 \\
Det. Noun Agreement &87 &90 &86 &\cellcolor[HTML]{E9E9E9}3 &\cellcolor[HTML]{E9E9E9}-1 &92 &93 &93 &88 &\cellcolor[HTML]{E9E9E9}1 &\cellcolor[HTML]{E9E9E9}1 &\cellcolor[HTML]{E9E9E9}-5 \\
Ellipsis &79 &79 &72 &\cellcolor[HTML]{E9E9E9}0 &\cellcolor[HTML]{E9E9E9}-7 &70 &71 &77 &70 &\cellcolor[HTML]{E9E9E9}1 &\cellcolor[HTML]{E9E9E9}7 &\cellcolor[HTML]{E9E9E9}-1 \\
Filler Gap &63 &70 &63 &\cellcolor[HTML]{E9E9E9}7 &\cellcolor[HTML]{E9E9E9}0 &69 &67 &74 &64 &\cellcolor[HTML]{E9E9E9}-2 &\cellcolor[HTML]{E9E9E9}5 &\cellcolor[HTML]{E9E9E9}-3 \\
Irregular Forms &76 &90 &86 &\cellcolor[HTML]{E9E9E9}14 &\cellcolor[HTML]{E9E9E9}10 &83 &92 &85 &84 &\cellcolor[HTML]{E9E9E9}9 &\cellcolor[HTML]{E9E9E9}2 &\cellcolor[HTML]{E9E9E9}-8 \\
Island Effects &44 &44 &37 &\cellcolor[HTML]{E9E9E9}0 &\cellcolor[HTML]{E9E9E9}-7 &49 &45 &52 &43 &\cellcolor[HTML]{E9E9E9}-4 &\cellcolor[HTML]{E9E9E9}3 &\cellcolor[HTML]{E9E9E9}-2 \\
NPI Licensing &58 &58 &55 &\cellcolor[HTML]{E9E9E9}0 &\cellcolor[HTML]{E9E9E9}-3 &55 &59 &68 &53 &\cellcolor[HTML]{E9E9E9}4 &\cellcolor[HTML]{E9E9E9}13 &\cellcolor[HTML]{E9E9E9}-6 \\
Quantifiers &73 &78 &73 &\cellcolor[HTML]{E9E9E9}5 &\cellcolor[HTML]{E9E9E9}0 &71 &68 &68 &71 &\cellcolor[HTML]{E9E9E9}-3 &\cellcolor[HTML]{E9E9E9}-3 &\cellcolor[HTML]{E9E9E9}3 \\
Subj. Verb Agreement &64 &70 &60 &\cellcolor[HTML]{E9E9E9}6 &\cellcolor[HTML]{E9E9E9}-4 &75 &75 &76 &66 &\cellcolor[HTML]{E9E9E9}0 &\cellcolor[HTML]{E9E9E9}1 &\cellcolor[HTML]{E9E9E9}-9 \\
Hypernym &50 &50 &50 &\cellcolor[HTML]{E9E9E9}0 &\cellcolor[HTML]{E9E9E9}0 &48 &48 &50 &49 &\cellcolor[HTML]{E9E9E9}0 &\cellcolor[HTML]{E9E9E9}2 &\cellcolor[HTML]{E9E9E9}1 \\
QA Congruence Easy &59 &56 &56 &\cellcolor[HTML]{E9E9E9}-3 &\cellcolor[HTML]{E9E9E9}-3 &64 &69 &66 &64 &\cellcolor[HTML]{E9E9E9}5 &\cellcolor[HTML]{E9E9E9}2 &\cellcolor[HTML]{E9E9E9}-5 \\
QA Congruence Tricky &38 &35 &35 &\cellcolor[HTML]{E9E9E9}-3 &\cellcolor[HTML]{E9E9E9}-3 &28 &34 &28 &28 &\cellcolor[HTML]{E9E9E9}6 &\cellcolor[HTML]{E9E9E9}0 &\cellcolor[HTML]{E9E9E9}-6 \\
Subject Aux Inversion &82 &78 &81 &\cellcolor[HTML]{E9E9E9}-4 &\cellcolor[HTML]{E9E9E9}-1 &70 &71 &76 &70 &\cellcolor[HTML]{E9E9E9}1 &\cellcolor[HTML]{E9E9E9}6 &\cellcolor[HTML]{E9E9E9}-1 \\
Turn Taking &67 &65 &55 &\cellcolor[HTML]{E9E9E9}-2 &\cellcolor[HTML]{E9E9E9}-12 &61 &59 &60 &61 &\cellcolor[HTML]{E9E9E9}-2 &\cellcolor[HTML]{E9E9E9}-1 &\cellcolor[HTML]{E9E9E9}2 \\
\cellcolor[HTML]{A8A8A8}Average &\cellcolor[HTML]{A8A8A8}66.5 &\cellcolor[HTML]{A8A8A8}67.9 &\cellcolor[HTML]{A8A8A8}63.6 &\cellcolor[HTML]{A8A8A8}1.4 &\cellcolor[HTML]{A8A8A8}-2.9 &\cellcolor[HTML]{A8A8A8}66.7 &\cellcolor[HTML]{A8A8A8}67.7 &\cellcolor[HTML]{A8A8A8}\textbf{69.1} &\cellcolor[HTML]{A8A8A8}64.5 &\cellcolor[HTML]{A8A8A8}0.9 &\cellcolor[HTML]{A8A8A8}2.4 &\cellcolor[HTML]{A8A8A8}-3.2
\\

\bottomrule
\end{tabular}
\caption{
BLiMP Results}\label{blimp_1}

\end{table*}

According to the BLiMP scores %
 in Table \ref{blimp_1}, within the \textit{Set A} models, the models incorporating hierarchical inductive bias (SF\(_{s1}\) and SF\(_{s2}\)) do not show consistent outperformance or underperformance in comparison to the baseline model TF\(_{base}\).

However, on average, the SF\(_{s1}\) model is on par with and occasionally outperforms the TF\(_{base}\) model. In particular, SF\(_{s1}\) excels in the following tests: Argument Structure, Determiner Noun Agreement, Filler Gap, Irregular Forms, Quantifiers, and Subj. Verb Agreement. Conversely, SF\(_{s1}\) underperforms the TF\(_{base}\) in the tasks of QA Congruence Easy, Subject Aux Inversion and Turn Taking. We hypothesize that this is because syntactic knowledge is helpful for the former list of tasks, but to a lesser degree for the latter, for example, Turn Taking, which focuses on knowledge of discourse and dialogue structure, in particular of referential properties of NPs, which is not reflected in the syntactic structure. A sample pair from this data set is \textit{"Should you quit?" -- "No, I shouldn't."} (good) versus \textit{"Should she quit?" --  "No, I shouldn't."} (bad). The negative and the positive data points have the same syntactic structure and the dependents are perfectly fine as argument fillers.

While the model with a parser network in-between the middle layers SF\(_{s2}\), underperforms TF\(_{base}\) on average, but interestingly it demonstrates a noteworthy improvement in the specific task of Irregular Forms. Remarkably, similar to SF\(_{s1}\), SF\(_{s2}\) significantly outperform TF\(_{base}\) in this particular task. The task of Irregular Forms involves aspects of lexical decisions but the syntax of course also plays a role.

Within the RoBERTa model variations in Set B, again the model with a parser network in-between the middle layers SR\(_{s2}\) fails to improve over the one with a parser network ahead of the encoder layers SR\(_{s1}\) in most of the tasks. It even gets worse with the upgrade in the number of convolution layers within the parser network at SR\(_{s2'}\). On the other hand, the upgrade in the number of convolution layers at SR\(_{s1'}\) shows also an upgrade in accuracies over SR\(_{s1}\). Generally, SR\(_{s1'}\) achieves the best results among all the investigated models on average.

Moreover, the Set B models exhibit improvements over Set A models in the tests of Binding, Det. Noun Agreement, Subject Verb Agreement, and QA Congruence Easy.

It is not so clear how to interpret the results of the two Question Answering (QA) Congruence tasks, where the baselines achieve only very low scores. %
 For the QA Congruence Easy task, which tests for detecting selectional preference violations on object fillers in answers (e.g., \textit{"What did you sell? - A chair."} (good) versus \textit{"What did you sell? - Sarah."} (bad)), knowing about the syntactic structure of the first sentence probably helps to apply selectional restrictions and thereby assessing the quality of the second as a possible reply. This might be the reason why we see an improvement in model performance in the SR models when adding implicit hierarchical information that reflects syntactic dependencies. The QA Congruence Tricky task is similar, except that the selectional preference that is violated in the negative data points does not refer to the direct object. Furthermore, the object is dropped in most examples and sometimes the (incorrect) argument filler would be a plausible direct object (e.g., \textit{"Who ate? - Sarah ate."} (good) versus \textit{"Who ate? - Pasta ate."} (bad)). This is why the task is tricky. In this context, it is important to keep in mind that our StructFormer models learn only unlabeled dependencies and therefore cannot distinguish between object and subject. This means that for \textit{Pasta ate}, a structure would be implicitly predicted where \textit{pasta}  is a dependent of \textit{ate}, which is perfectly fine semantically (as a direct object). This might be a reason why the structformer models struggle with this test and partly lead to a decrease in the performance, compared to our baseline, since the unlabeled dependency tree actually licenses the negative data points.

\subsection{SuperGLUE}

SuperGLUE consists of eleven diverse tasks (see Appendix~\ref{appendix: evaluation_metrics}) which evaluate various performance aspects. These tasks include sentiment analysis, linguistic acceptability judgments, entailment detection, and semantic similarity evaluations of words within contexts, among others \cite{10.5555/3454287.3454581}.

\begin{table*}[ttp]\centering
\scriptsize
\begin{tabular}{p{1.5cm}rrrrrrrrrrrr}
\toprule
& 
\multicolumn{5}{c}{\textbf{Set A}}    
& &
\multicolumn{5}{c}{\textbf{Set B}}     
& 
\\\midrule

& \textbf{TF\(_{base}\)} 
& \textbf{SF\(_{s1}\)} 
& \textbf{SF\(_{s2}\)} 
&\cellcolor[HTML]{E9E9E9} \textbf{\(\Delta_{SF_{s1}}\) }
&\cellcolor[HTML]{E9E9E9} \textbf{\(\Delta_{SF_{s2}}\) } 
& \textbf{SR\(_{s1}\)} 
& \textbf{SR\(_{s2}\)} 
&\textbf{SR\(_{s1'}\)} 
&\textbf{SR\(_{s2'}\)} 
&\cellcolor[HTML]{E9E9E9}\textbf{ \(\Delta_{SR_{s12}} \) } 
&\cellcolor[HTML]{E9E9E9}\textbf{ \(\Delta_{SR_{s1'}} \) } 
&\cellcolor[HTML]{E9E9E9}\textbf{ \(\Delta_{SR_{s2'}} \) } 
\\\midrule

BoolQ &63 &61 &62 &\cellcolor[HTML]{E9E9E9}-2 &\cellcolor[HTML]{E9E9E9}-1 &66 &66 &64 &65 &\cellcolor[HTML]{E9E9E9}0 &\cellcolor[HTML]{E9E9E9}-2 &\cellcolor[HTML]{E9E9E9}-1 \\
COLA (MCC) &0.16 &0.19 &0.14 &\cellcolor[HTML]{E9E9E9} --- &\cellcolor[HTML]{E9E9E9}--- &0.23 &0.23 &0.19 &0.26 &\cellcolor[HTML]{E9E9E9}--- &\cellcolor[HTML]{E9E9E9}--- &\cellcolor[HTML]{E9E9E9}--- \\
MNLI &71 &71 &70 &\cellcolor[HTML]{E9E9E9}0 &\cellcolor[HTML]{E9E9E9}-1 &72 &72 &69 &72 &\cellcolor[HTML]{E9E9E9}0 &\cellcolor[HTML]{E9E9E9}-3 &\cellcolor[HTML]{E9E9E9}0 \\
MNLI-MM &72 &73 &72 &\cellcolor[HTML]{E9E9E9}1 &\cellcolor[HTML]{E9E9E9}0 &73 &73 &70 &73 &\cellcolor[HTML]{E9E9E9}0 &\cellcolor[HTML]{E9E9E9}-3 &\cellcolor[HTML]{E9E9E9}0 \\
MRPC (F1) &75 &75 &79 &\cellcolor[HTML]{E9E9E9}0 &\cellcolor[HTML]{E9E9E9}4 &76 &81 &77 &75 &\cellcolor[HTML]{E9E9E9}5 &\cellcolor[HTML]{E9E9E9}1 &\cellcolor[HTML]{E9E9E9}-6 \\
MultiRC &61 &58 &62 &\cellcolor[HTML]{E9E9E9}-3 &\cellcolor[HTML]{E9E9E9}1 &62 &59 &59 &54 &\cellcolor[HTML]{E9E9E9}-3 &\cellcolor[HTML]{E9E9E9}-3 &\cellcolor[HTML]{E9E9E9}-5 \\
QNLI &81 &77 &78 &\cellcolor[HTML]{E9E9E9}-4 &\cellcolor[HTML]{E9E9E9}-3 &71 &72 &66 &74 &\cellcolor[HTML]{E9E9E9}1 &\cellcolor[HTML]{E9E9E9}-5 &\cellcolor[HTML]{E9E9E9}2 \\
QQP (F1) &81 &82 &81 &\cellcolor[HTML]{E9E9E9}1 &\cellcolor[HTML]{E9E9E9}0 &82 &82 &80 &81 &\cellcolor[HTML]{E9E9E9}0 &\cellcolor[HTML]{E9E9E9}-2 &\cellcolor[HTML]{E9E9E9}-1 \\
RTE &48 &42 &47 &\cellcolor[HTML]{E9E9E9}-6 &\cellcolor[HTML]{E9E9E9}-1 &46 &57 &53 &56 &\cellcolor[HTML]{E9E9E9}11 &\cellcolor[HTML]{E9E9E9}7 &\cellcolor[HTML]{E9E9E9}-1 \\
SST2 &87 &85 &82 &\cellcolor[HTML]{E9E9E9}-2 &\cellcolor[HTML]{E9E9E9}-5 &87 &82 &86 &83 &\cellcolor[HTML]{E9E9E9}-5 &\cellcolor[HTML]{E9E9E9}-1 &\cellcolor[HTML]{E9E9E9}1 \\
WSC &61 &61 &61 &\cellcolor[HTML]{E9E9E9}0 &\cellcolor[HTML]{E9E9E9}0 &61 &59 &61 &61 &\cellcolor[HTML]{E9E9E9}-2 &\cellcolor[HTML]{E9E9E9}0 &\cellcolor[HTML]{E9E9E9}2 
\\
\bottomrule
\end{tabular}
\caption{(Super)GLUE Results. Values are not aggregated across each model due to the presence of different metrics (Accuracy, F1 score, and MCC)}\label{glue_1}
\end{table*}

The scores (see Table \ref{glue_1}) in most of the tasks fall in a narrow range across all the investigated models. The incorporation of hierarchical inductive bias does not show clear improvements in most of the tasks. A noticeable result that is observed for the models with a parser network within the middle layers \textit{(s2)} is the result of the MRPC task, where \textit{s2} models consistently outperform the \textit{s1} models in both sets for this particular task. The upgrade in the number of convolution layers also does not show a clear improvement in most of the tasks for both SR\(_{s1'}\) and SR\(_{s2'}\) models.

Notably, in the case of the WSC task, we observe that all models' predictions heavily favored one specific class. This raises concerns about the success of the finetuning process for this particular task.

\subsection{MSGS}

The MSGS tasks, listed in Appendix~\ref{appendix: evaluation_metrics}, were introduced by the shared task as held-out tests specifically designed to evaluate generalization capabilities. Detailed information and further insights about these tasks are expected to be disclosed in an upcoming publication. MSGS tasks are measured using the Matthews correlation coefficient (MCC). MCC is used in machine learning as a measure of the quality of binary (two-class) classifications, introduced by \citet{MATTHEWS1975442}

The MSGS results (Table \ref{msgs_1}), resemble to %
 the SuperGLUE results. The models incorporating hierarchical inductive bias show contradicting behavior across the different tasks. While for some tasks e.g Control Raising (Control), Relative Position (Control), and Syntactic Category (Relative Position), SF\(_{s1}\) and SF\(_{s2}\) are strengthening the correlation in comparison to the baseline model, but with other tasks e.g Lexical Content (Control), Main Verb (Lexical Content) and Syntactic Category (Lexical Content), SF\(_{s1}\) and SF\(_{s2}\) are shown weakening the correlation.

\begin{table*}[!htp]\centering
\scriptsize
\begin{tabular}{lrrr|rrrr}
\toprule
& 
& Set A
&
& 
& Set B
& 
&
\\\midrule

& \textbf{TF\(_{base}\)} 
& \textbf{SF\(_{s1}\)} 
& \textbf{SF\(_{s2}\)} 
& \textbf{SR\(_{s1}\)} 
& \textbf{SR\(_{s2}\)} 
&\textbf{SR\(_{s1'}\)} 
&\textbf{SR\(_{s2'}\)} 
\\\midrule

Control Raising (Control) &0.54 &0.56 &0.69 &0.57 &0.56 &0.69 &0.56 \\
Control Raising (Lexical Content) &-0.45 &-0.04 &-0.02 &-0.03 &-0.07 &-0.36 &-0.14 \\
Control Raising (Relative Position) &-0.94 &-0.89 &-0.92 &-1.00 &-0.98 &-0.77 &-0.98 \\
Lexical Content (Control) &1.00 &0.88 &0.6 &1.00 &0.98 &1.00 &0.78 \\
Main Verb (Control) &0.93 &0.96 &0.84 &0.85 &0.98 &0.96 &0.98 \\
Main Verb (Lexical Content) &-1.00 &-0.79 &-0.84 &-1.00 &-1.00 &-0.99 &-1.00 \\
Main Verb (Relative Position) &-0.87 &-0.78 &-0.89 &-0.98 &-0.93 &-0.83 &-0.95 \\
Relative Position (Control) &0.67 &0.81 &0.78 &0.86 &0.95 &0.97 &1.00 \\
Syntactic Category (Control) &0.62 &0.23 &0.47 &0.80 &0.73 &0.66 &0.87 \\
Syntactic Category (Lexical Content) &-0.61 &-0.17 &-0.17 &-0.42 &-0.59 &-0.26 &-0.76 \\
Syntactic Category (Relative Position) &-0.32 &-0.57 &-0.44 &-0.47 &-0.47 &-0.63 &-0.52 \\
\bottomrule
\end{tabular}
\caption{MSGS Results}\label{msgs_1}
\end{table*}

\subsection{Aggregation}

Indeed, analyzing the performance changes across 39 tasks for 7 different models is a complex process. To simplify the assessment and present a concise summary of each model's overall performance, we report an aggregate score of all the 39 scores for each model (Table~\ref{db_scores}). This aggregation approach was internally computed by the shared task submission platform to represent each model with a single score, providing a more straightforward evaluation of the overall performance. Subsequently, we select the model with the best aggregate score SR\(_{s1'}\) to represent our submission in the shared task.

\begin{table*}[!htp]\centering
\scriptsize
\begin{tabular}{lrrr|rrrr}
\toprule
& 
& Set A
&
& 
& Set B
& 
&
\\\midrule

& \textbf{TF\(_{base}\)} 
& \textbf{SF\(_{s1}\)} 
& \textbf{SF\(_{s2}\)} 
& \textbf{SR\(_{s1}\)} 
& \textbf{SR\(_{s2}\)} 
&\textbf{SR\(_{s1'}\)} 
&\textbf{SR\(_{s2'}\)} 
\\\midrule

Aggregate Score &0.52 &0.53 &0.52 &0.53 &0.54 &\textbf{0.55} &0.52 \\
\bottomrule
\end{tabular}
\caption{Shared Task Leaderboard Results}\label{db_scores}
\end{table*}

\section{Discussion}

Although the evaluation pipeline of the shared task was meticulously designed to encompass a comprehensive analysis of pretrained LMs, covering aspects of efficiency, applied NLP standards, cognitive science, linguistics, and language acquisition \cite{warstadt-et-al-2023-babylm}, it was discussed in \citet{warstadt-etal-2020-blimp-benchmark} that some tasks that involve semantic phenomena such as Island Effects and NPI Licensing are very difficult for LMs in general. %
 Consequently, the consistently low performance observed across all models on these tests can be attributed to this matter. As a result, we refrain from considering the aggregate score as a single definitive metric for representing how a model's performance compares to another. Instead, we advocate for a thorough investigation of individual tests while considering the test's objectives, dataset, and evaluation strategy.

Overall, the models incorporating hierarchical inductive bias did not show significant improvement in the scores of the BabyLM evaluation tasks, however, some exceptions of the evaluation tasks that show improvements in terms of scores when using the \textit{structformer} and \textit{structroberta} models, encourage a deeper investigation for patterns in the outputs predictions that might lead to a different conclusion. Namely, the tasks that we think are worth more investigation are: \textit{Argument Structure, Determiner Noun Agreement, Filler Gap, Irregular Forms, Quantifiers, Subj.
Verb Agreement, Control Raising (Control), Relative Position (Control) and Syntactic Category (Relative Position)}.

Contrary to our expectations, the modification of placing the parser in-between the middle attention layers has not %
 demonstrated notable improvements but rather a decline in performance compared to the models with the parser placed right after the input embedding layer. We can only speculate about why this is so. It might be that it is an advantage to push the model very early towards identifying structural relations between words. More precisely to do so at a stage where the contributions of the single tokens are still separated from each other. The parsing network placed between the middle layers acts at a moment where single token contributions are already blurred.

To understand the effect of placing the parser network within the middle layers, we propose probing the layers of the Front and Rear modules and comparing them to the corresponding layers in the model where the parser network is placed ahead of the attention layers. Such a comparative analysis can provide valuable insights and either support or contradict our hypothesis regarding the learning of syntactic features in the middle layers of transformer models.

Regarding the aim of achieving competitive scores on the shared task challenge, the best score we could get was from the model \textit{structroberta\(_{s1'}\)}, this model is an upscaling of the \textit{structroberta\(_{s1}\)}.

\section{Conclusion}

In this paper, we extend the work of \citet{shen-etal-2021-structformer} to explore the capabilities of the StructFormer architecture as an example of employing hierarchical bias in addressing the challenges posed by relatively small LLM pretraining datasets. 
Furthermore, we modify the StructFormer architecture to examine whether integrating the hierarchical bias within the middle attention layers leads to performance improvements. 
To accomplish these objectives, we pretrain seven model variants using the same dataset and configuration settings. We evaluate these models on 39 different tasks. The evaluation outcomes reveal varying behavior across the models, exhibiting inconsistencies in performance. We could not show strong evidence that models incorporating hierarchical bias are performing better in the context of this shared task, nor could we show practical evidence for the claim that syntactic information is better represented in the middle attention layers within the scope of our experiment. We have noted substantial enhancements in certain tasks when models incorporate hierarchical bias in their architectural designs. Nonetheless, to ensure the reliability of our findings and to eliminate potential confounding factors related to the varying number of parameters in each model, as well as the distinct objectives and complexities of individual tasks, we intend to carry out an in-depth analysis of each model's performance on a task-by-task basis.

\section*{Acknowledgements}

We thank the authors of the StuctFormer model \cite{shen-etal-2021-structformer} for providing their implementation, which played an important role in the completion of this work. 
Additionally, we acknowledge the invaluable support received from the BabyLM shared task organizers, who provided the datasets, evaluation pipeline, and codes for pretraining and finetuning LMs. 
Their contributions enabled us to conduct a comprehensive and successful study. Furthermore, we are grateful for the comments of our reviewers that helped improve the paper.
Lastly, we thank Hassan Sajjad and Younes Samih for fruitful discussions on hierarchical information in language models.

\bibliography{babylm_emnlp2023}
\bibliographystyle{acl_natbib}

\section*{Appendix} 
\appendix

\section{Evaluation Tasks}
\label{appendix: evaluation_metrics}

\textbf{BLiMP}
\begin{enumerate}
    \item Anaphor Agreement
    \item    Argument Structure
    \item    Binding
    \item    Control Raising
    \item    Determiner Noun Agreement
    \item    Ellipsis 
    \item    Filler Gap 
    \item    Irregular Forms 
    \item    Island Effects 
    \item    Negative Polarity Items NPI Licensing
    \item    Quantifiers
    \item    Subject Verb Agreement
    \item    Hypernym
    \item    QA Congruence Easy
    \item    QA Congruence Tricky
    \item    Subject Aux Inversion 
    \item    Turn Taking
\end{enumerate}

\textbf{SuperGLUE}
\begin{enumerate}
\setcounter{enumi}{17}
\item Corpus of Linguistic Acceptability CoLA (MMC)
\item Stanford Sentiment Treebank SST-2
\item Microsoft Research Paraphrase Corpus MRPC (F1)
\item Quora Question Pairs QQP (F1) 
\item MultiNLI Matched MNLI
\item MultiNLI Mismatched MNLI-mm
\item Question NLI QNLI
\item Recognizing Textual Entailment RTE
\item Boolean Questions BoolQ
\item Multi-Sentence Reading Comprehension MultiRC
\item Winograd Schema Challenge WSC
\end{enumerate}

\textbf{MSGS}
\begin{enumerate}
\setcounter{enumi}{28}
\item Main Verb (Control)
\item Control Raising (Control)
\item Syntactic Category (Control)
\item Relative Position (Control) 
\item Lexical Content The (Control)
\item Main Verb Lexical Content The
\item Main Verb Relative Token Position
\item Control Raising Lexical Content The
\item Control Raising Relative Token Position
\item Syntactic Category Lexical Content The
\item Syntactic Category Relative Position
\end{enumerate}

\section{Hyperparameters and Models Configurations}
\label{appendix: hyperparams}
In Table \ref{table: n_params}, we report the number of trainable parameters per model. In Table \ref{table: hyps}, we report all the important hyperparameters values for all our pretraining and finetuning experiments. Also, we report the main configuration settings for all our models. Unless specified otherwise, these values were used across all models. 

\begin{table}[!htp]\centering
\scriptsize
\begin{tabular}{lrr}\toprule
Model & \# of trainable parameters \\\midrule
transformer-base & 110M \\
structformer\(_{s1}\) & 133M \\
structformer\(_{s2}\) & 133M \\
structroberta\(_{s1}\) & 133M \\
structroberta\(_{s2}\) & 133M \\
structroberta\(_{s1'}\) & 144M \\
structroberta\(_{s2'}\) & 144M \\
\bottomrule
\end{tabular}
\caption{Number of trainable parameters per model}\label{table: n_params}
\end{table}

\begin{table}[!htp]\centering
\scriptsize
\begin{tabular}{lrr}\toprule
Training Hyperparameters & \\\midrule
Batch size &96 \\
Sequence Length &128 \\
Optimizer &AdamW \\
Weight Decay &0.1 \\
Learning Rate (Linear) &1e-4 \\
Max Steps &62K \\
Masking Probability &0.15 \\
& \\\midrule
Finetuning Hyperparameters & \\\midrule
Initial learning rate &5e-5 \\
Batch size &120 \\
Maximum epochs &10 \\
Evaluate every (steps) &400 \\
Patience &5 \\
Random seed &12 \\
& \\\midrule
Models configurations & \\\midrule
Number of Attention Heads &12 \\
Number of Attention Layers &12 \\
Embeddings (Hidden) Size &768 \\
FFN inner hidden size &3072 \\
Attention Dropout &0.1 \\
Front Attention Layers (where applicable) &4 \\
Rear Attention Layers (where applicable) &8 \\
Parser Convolution Layers (where applicable) &4 \\
Convolution Kernel Size &9 \\
\bottomrule
\end{tabular}
\caption{Pretraining and Finetuning Hyperparameters, and Models Configurations Settings}\label{table: hyps}
\end{table}

\end{document}